%% file: root.tex
\DocumentMetadata{
  pdfversion=2.0,pdfstandard=ua-2,
  testphase={phase-III,firstaid,math,title}
}
\documentclass[sigconf,screen]{acmart-tagged}

\makeatletter
\AtBeginDocument{%
  \DeclareCaptionLabelFormat{bold}{\textbf{#1~#2}}
  \captionsetup{labelformat=bold, labelfont={bf}}
  
}
\makeatother

\Description{HRI paper}

\AtBeginDocument{%
  }

\setcopyright{cc}
\setcctype{by}
\acmDOI{10.1145/3757279.3785635}
\acmYear{2026}
\copyrightyear{2026}
\acmISBN{979-8-4007-2128-1/2026/03}
\acmConference[HRI '26]{Proceedings of the 21st ACM/IEEE International Conference on Human-Robot Interaction}{March 16--19, 2026}{Edinburgh, Scotland, UK}
\acmBooktitle{Proceedings of the 21st ACM/IEEE International Conference on Human-Robot Interaction (HRI '26), March 16--19, 2026, Edinburgh, Scotland, UK}
\received{2025-09-30}
\received[accepted]{2025-12-01}

\usepackage{siunitx}
\usepackage{balance}

\makeatletter
\let\orig@includegraphics\includegraphics
\renewcommand{\includegraphics}[2][]{%
  \in@{doclicense-CC}{#2}%
  \ifin@
    \orig@includegraphics[#1]{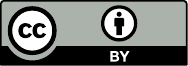}%
  \else
    \orig@includegraphics[#1]{#2}%
  \fi
}
\makeatother

\begin{document}

\title{Context-Aware Generation and Modulation of Expressive Motion Behavior using Multimodal Foundation Models}

\author{Till Hielscher}
\affiliation{%
  \institution{University of Stuttgart}
  \city{Stuttgart}
  \country{Germany}
}
\email{till.hielscher@ki.uni-stuttgart.de}

\author{Fabio Scaparro}
\affiliation{%
  \institution{University of Stuttgart}
  \city{Stuttgart}
  \country{Germany}
}
\email{fabio.scaparro@ki.uni-stuttgart.de}

\author{Kai O. Arras}
\affiliation{%
  \institution{University of Stuttgart}
  \city{Stuttgart}
  \country{Germany}
}
\email{kai.arras@ki.uni-stuttgart.de}

\renewcommand{\shortauthors}{Hielscher et al.}

\begin{abstract}
  \input{sections/abstract}

\end{abstract}

\ccsdesc[500]{Computing methodologies~Robotic planning}

\keywords{Expressive robot motion behavior, Context awareness, Multimodal foundation models, Knowledge distillation}

\maketitle

\input{sections/introduction}

\input{sections/related_work}

\input{sections/method}

\input{sections/experiments}

\input{sections/results}

\input{sections/conclusion}

\begin{acks}
The authors thank the International Max Planck Research School for Intelligent Systems (IMPRS-IS)
for supporting Till Hielscher.
\end{acks}

\balance

\bibliographystyle{ACM-Reference-Format}
\bibliography{bibliography}

\end{document}

%% file: sections/abstract.tex
Expressive robot motion positively impacts human-robot interaction by improving user engagement, likability, or task performance. We present a novel approach to automatically generate and modulate complex, expressive full-body gesture sequences from a multimodal (text, audio, video) context description. Based on a unified mathematical implementation of the Principles of Animation using Dynamic Movement Primitives, we use multimodal foundation models to generate such sequences along with parametric motion variations that are highly context-aware. This method extends the state of the art in terms of flexibility and generality by being interpretable, composable, and working across different robot morphologies. Moreover, we integrate the system into a continuous control framework and leverage knowledge distillation to learn a much smaller model, significantly improving token efficiency and system latency. Results from a user study with a human-like platform indicate that participants judged our system’s motions to be better aligned with the interaction context than motions produced under a non-modulated or API-level condition. We also demonstrate how the system generates expressive motion for robots with different kinematics, showcasing its versatility. Paper webpage: \href{https://gen-mod-expressive-motion-behavior.github.io/}{https://gen-mod-expressive-motion-behavior.github.io/}.

%% file: sections/introduction.tex
\section{Introduction}
    \begin{figure}[t]
        \centering
        \includegraphics[width=0.99\linewidth, alt={Showcase of the timeline with corresponding gestures and modulations as well as the executed motion on the Pepper robot. The audio triggers generation based on an extended context description. The generated motion behavior is appended to the timeline and executed.}]{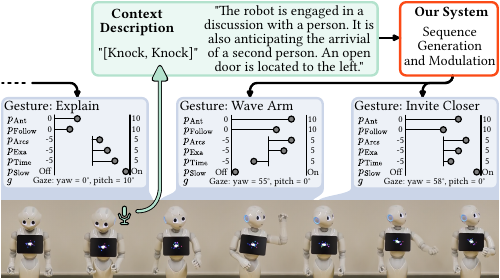}
        \vspace{-3mm}
        \caption{\textbf{Given a multi-modal context description (green: audio and text), our method composes modulated gestures (blue) into expressive, context-aware body motion. In this scenario, a robot interacting with a person detects a knocking sound, pauses the interaction, turns toward the door, and greets the arriving person in a neutral way. If the context called for a different reaction (e.g., joy at a long-awaited reunion), the result would be a corresponding modulation, such as exaggerated greeting.}}
        \Description{Showcase of the timeline with corresponding gestures and modulations as well as the executed motion on the Pepper robot. The audio triggers generation based on an extended context description. The generated motion behavior is appended to the timeline and executed.}
        \label{fig:title}
        \vspace{-3mm}
    \end{figure}
    
    Non-verbal communication via gaze, gestures, or facial expressions is a crucial component of socially intelligent behavior for humans and robots.
    Robot-performed gestures are often seen as a defining property of social robots \cite{de2023design}, and by using their physical embodiment and presence, they are perceived more positively and persuasive than, for example, interactive virtual agents \cite{li2015benefit}. Therefore, the design and use of expressive robotic body motion are critical for achieving natural and effective communication with socially interactive robots.
    Key questions include how movements are shaped, selected, and timed, the extent of their expression, and their integration with speech, context, and the course of interaction. 
    This paper tackles several of the above questions, namely how body motion can be represented to facilitate learning, execution, and modulation -- i.e., context-sensitive variations of the same motion --, and how basic gestures can be automatically selected, timed, and varied such that interlocutors perceive the robot’s behavior as aligned with the interaction context.    

    The variety of expressive human body motion appears to be open-ended. One way to tackle this complexity practically is categorizing gestures into communicative functions such as deictic, iconic, or symbolic, and reusing standardized sets of pre-recorded gestures \cite{tan2021proposedgestures}. By providing a small set of exemplars for each function, a robot can cover many real-world situations without the need to generate joint-level trajectories on the fly.
    We build on such motion datasets and adopt recent work \cite{hielscher2025interactive} that employs Dynamic Movement Primitives (DMPs) to represent, modulate, and compose basic gestures into expressive body language. Originally developed for motor skill learning from imitation or demonstration, e.g., for robot manipulation, DMPs offer properties highly relevant for this task. As shown in \cite{hielscher2025interactive}, they provide a unified implementation of the Principles of Animation and enable modulation of key principles through a low-dimensional, interpretable parameter set. This allows animators -- and large language models -- to stylize full-body motion along the well-established dimensions of the Principles of Animation across different kinematic configurations. They are also online adaptive in the sense that gestures are parametrized by a static or dynamic goal in space towards which gestures are executed.
    
    However, unlike \cite{hielscher2025interactive}, which was limited to the ``how'' in generating expressive body language, we focus on the ``what'' and ``when'' and automate the generation process given the multimodal description of a context. To this end, we integrate multimodal foundation models into a continuous control framework, leveraging their extensive world knowledge to explore novel possibilities for social robot motion design. This also extends prior work that used only large language models and simpler motion representations \cite{mahadevan2024generative}. Specifically, our contributions are:
    \begin{itemize}
        \item A system for automated context-aware modulation and composition of basic gestures, represented as Dynamic Movement Primitives (DMPs), into expressive body language. The method is robot-agnostic with respect to kinematics and integrates multimodal context inputs, including text, image, and audio.
        \item A method that applies knowledge distillation, a technique to transfer the knowledge of a large, computationally expensive foundation model to a smaller, more efficient model. The results are reduced inference cost and significantly smaller system latencies.
        \item A user study with a human-like robot that confirms the validity of the approach over a non-modulated and API-level motion condition, as well as qualitative experiments that demonstrate the system's ability to generate expressive body language across robot morphologies using a manipulator and a four-legged robot. 
    \end{itemize}

%% file: sections/related_work.tex
\section{Related Work}
    \subsection{Expressive Motion Behavior Generation}
        Prior work has extensively explored the generation and use of gestures and expressive body language in HRI, see \cite{de2023design} for a survey. The literature can be divided into planning and generation methods. The former denotes the high-level task of selecting and composing basic movements into expressive motion behaviors, the latter is the lower-level task of synthesizing, learning, or executing basic movements, e.g., at the joint trajectory level.
        
        \textbf{High-Level Motion Behavior Generation:}
            Early works introduce methods for high-level behavior generation based on manually designed, rule-based state transition systems.
            Kato et al. \cite{kato2015may} point out that such systems require extensive understanding and classification of human activities, behaviors, and other input modalities for valid and reasonable state transitions.
            In \cite{buchina2016design}, the authors focus on the ease of use of such systems, while later works also deal with the formal verification of expressive rule-based behaviors \cite{porfirio2018authoring} or the application of a manually designed behavior system, e.g., for the Pepper robot \cite{leonardi2019trigger}.
        
            As the manual design of rule-based behaviors is tedious and counterintuitive, follow-up works focus on semi-automatic and automated creation of behavioral decision systems.
            The authors of \cite{porfirio2019bodystorming} synthesize formalisms from traces of instruction, and later provide modification options \cite{porfirio2020transforming} and user-friendly instruction authoring \cite{porfirio2023sketching}.        
            In order to handle the complexity of multi-modal context inputs automatically, the method proposed in \cite{huang2014learning} fits a Bayesian network to model multi-modal behaviors.
            Recent works utilize deep neural networks with reinforcement learning \cite{hussain2022training} and time-series information \cite{murray2022learning} to trigger gestures during interaction.
            With the goal of general context understanding, some works also utilize Large Language Models (LLMs) to guide the behavior of the robot.
            In \cite{kumar2023words}, the authors introduce a framework that first derives motion policies from a large-scale human motion generation model, and then employs a language model to determine whether a new task aligns with an existing policy and to refine it accordingly, thereby enabling efficient policy reuse and natural language-based specification of complex motion behaviors.
            Based on scenario descriptions, Mahadevan et al. \cite{mahadevan2024generative} generate API-level programs with a multi-stage LLM system for context understanding and code generation. Only by providing robot-specific information and a library of available API calls the system can generate expressive behaviors for various robots.
            Similarly, LLMs can enable the relation of a target expression with the available hardware using motion parameters. \cite{roy2025gpt} shows results for a quadruped robot by providing the available degrees of freedom as parameters of the robot, along with descriptions that describe a possible effect.

            For high-level behavior generation, we utilize general context understanding to alleviate the limitations of manual and rule-based methods in terms of flexibility. However, instead of simple calls to the robots' actuators \cite{mahadevan2024generative} or limitation to specific morphologies \cite{roy2025gpt}, we generate sequences of complex whole-body gestures based on a pre-recorded library represented as Dynamic Movement Primitives \cite{schaal2002movement}. This enables the generation of cross-morphology context-aligned expressive motion, while providing the additional feature of risk management of gestures that might be perceived as political, obscene, or intensive by simply excluding them from the library.

        \textbf{Low-Level Motion Representations:}
            In robotics, the challenge of low-level motion generation mirrors that of character animation, where the design of trajectories must balance physical feasibility with expressive qualities that communicate intent, emotion, or style to a human observer. Important works \cite{lasseter1987principles}, and tools like Aldebaran Choreographe utilize trajectory shaping with keyframe interpolation using splines or polynomials. The method provides considerable flexibility at the cost of laborious manual design. Later, methods for motion representation like Dynamic Movement Primitives \cite{schaal2002movement} for imitation of demonstrated motion or the probabilistic extension \cite{paraschos2013probabilistic} are proposed. While offering smooth execution capabilities in their basic forms, they lack flexibility towards expressive robot motion generation.
            
            By mapping the latent space of a variational autoencoder to the Valence Arousal Dominance (VAD) space, the method proposed in \cite{marmpena2019generating} can adjust input motion to align with targeted VAD states. The method is limited to a single kinematic structure on which the model is trained.
            In \cite{oh2025mowa}, motion is adjusted by modulating along the Principles of Animation in the latent space of an encoder-decoder structured model for human avatars.
            The method proposed in \cite{hielscher2025interactive} also relies on the representation space of the Principles of Animation, but encodes them in a physics-based model, which is decoupled from learned kinematic structures and enables online modulation.
            The authors of \cite{alegre2025amor} show that also reinforcement learning based systems can be trained for behavioral adjustments based on multi-objective reward functions. The model uses reward weights that can be tuned to modulate motion online.
            
            Towards lively and autonomous non-verbal interaction, Christen et al. \cite{christen2025autonomous} propose a diffusion-based system that generates controls conditioned on a nearby human's pose and a target expressivity. The system is trained with expert data for a single robot.
            Some works also directly leverage LLMs to either adapt \cite{yu2024towards} or generate \cite{huang2025emotion} sequences of joint states to enable context-dependent expressivity. In both works, the model is provided with an understanding of the human's kinematic structure with differing levels of detail.
            
            Compared to these works, which lack dynamic consistency \cite{huang2025emotion}, modulate with limited flexibility \cite{marmpena2019generating}, or use latent space grounding \cite{oh2025mowa, alegre2025amor}, our method combines context-aware expressive adaptation with physically grounded motion execution and modulation. To this end, we extend our past work \cite{hielscher2025interactive} for fully automatic, online adaptive modulation parametrization.
        
        \textbf{Integrated Continuous Control:}
            Frameworks for continuous planning and control enable the deployment of robots for interaction by ensuring both timely responses to cues in active scenarios \cite{shiwa2008quickly} and lifelike behavior when idle \cite{asselborn2017keep}.

            Generally, and as shown in prior works, both \cite{lu2024implementation} and \cite{chen2019online} propose methods that combine a selection of perception and cognition models with action planning to continuously handle user inputs and react when necessary. Oleinikov et al. \cite{oleinikov2021safety} further introduce a hierarchical controller that operates at different frequencies, allowing a robotic arm to incorporate contextual human interaction information into motion planning at both high and low levels.
            Current works face ongoing challenges in combining general context understanding, flexibility, and fast reaction to achieve continuous control with expressive motion behavior. Our method addresses this by picking up the hierarchical control idea and condensing modularity with general understanding.

    \subsection{Co-Speech Gestures and Motion}
        Although our work does not address Co-Speech motion generation, we review related methods as they tackle similar challenges, especially with respect to selection and timing of gestures to enable synchronized behavior.

        Co-speech gestures can be coordinated with speech by either adapting the gestures to the spoken content or adapting the speech to the planned gestures \cite{de2023design}. Adapting gestures to speech involves timing motions to align with specific words, which can be done manually when the utterance is predetermined \cite{willemsen2018context}. More precise alignment can be achieved by pre-rendering speech to calculate word durations and combining this information with the robot's motor capabilities \cite{ng2010synchronized, shi2010easy}. Contextual factors such as accented syllables can be factored in to influence timing \cite{mlakar2013tts}. The alignment of speech with gestures can be realized by matching relevant speech units with related gesture units \cite{kipp2007towards}. 
        As reported in \cite{nyatsanga2023comprehensive}, learning-based methods trained for human avatars can leverage available data to generate co-speech gestures. These can incorporate multimodal temporal \cite{chiu2015predicting}, semantic \cite{yi2023generating, kucherenko2022multimodal}, or spatial grounding \cite{deichler2024incorporating}.

        For robotics applications, this positions our work to share similar goals with co-speech motion research, namely, context-aware generation of expressive behavior. While we do not focus on synchronizing multiple output modalities, the temporal grounding in the phase variable of the DMP could enable such coordination.

%% file: sections/method.tex
\section{Method}
    \begin{figure*}[t]
        \centering
        \includegraphics[width=0.99\linewidth, alt={An overview figure for the quantitative results of the experiments, which is split into three parts and visualizes the results given in the results sections.}]{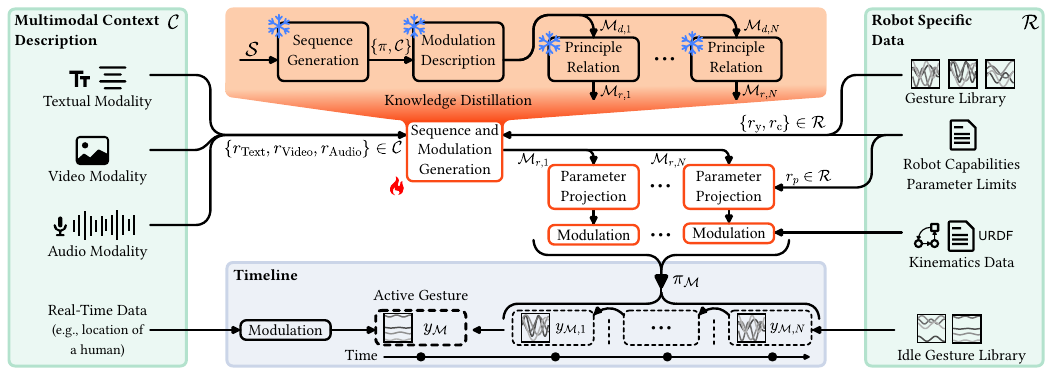}
        \vspace{-3mm}
        \caption{\textbf{Overview of our system: Given multimodal context descriptions and robot-specific parameters, our approach generates sequences of modulated gestures using a distilled multimodal foundation model. A reactive timeline-based controller manages continuous motion and perceived context changes from sensory data to modulate the active gesture. See text for details.}}
        \Description{Individual sections of the figure showcase different parts of the system as well as their connection. This is a visual representation of the aspects described in the method section.}
        \label{fig:system_overview}
    \end{figure*}
    In this section, we present our approach for the automatic generation of expressive, context-sensitive motion behavior. First, we describe the method for generating motion sequences that capture expressive behavior. Then, we detail how these sequences are modulated to ensure that the executed movements convey expressivity aligned with the context. Finally, we explain the integration of these components into a flexible and efficient framework that supports continuous operation. See \autoref{fig:title} and \autoref{fig:sequences} for example results of our method and \autoref{fig:system_overview} for the overview of the system.
    
    Interaction context is inherently multimodal, comprising a variety of spatial, social, and environmental cues that together shape how a robot's response is perceived as normative or appropriate. To achieve high levels of context sensitivity, robotic systems must integrate such cues in the form of verbal, auditory, and visual stimuli. %
    Thus, we define the input to our system to be a multimodal context description using text, audio, and video, as well as sensory inputs that arrive in real time. We denote this as the robot's interaction context $\mathcal{C} = \{r_{\text{Text}}, r_{\text{Video}}, r_{\text{Audio}}, r_{\text{RT}}\}$. Real-time inputs include, for example, spatial positions from a visual person tracker or object detector.
    
    \subsection{Sequence Generation} \label{sec:what}
        Context alignment critically depends on the choice of actions, their execution order, style, and timing. The literature suggests that efficient interaction is often based on a set of essential movements and gestures \cite{holler2022visual}. In line with these findings, our method generates motion sequences from pre-learned robotic skills that represent appropriate actions for various interaction scenarios. The main advantage of this approach is the utilization of complex, whole-body motions that already possess inherent communicative meaning while abstracting the underlying robot kinematics. This enables robots with arbitrary morphologies to utilize complex interaction gestures for effective communication and %
        bridges the embodiment gap by providing a hardware-agnostic mapping between generative models and physical robot execution.
        
        Given an interaction context $\mathcal{C}$, we address the open-ended problem of generating context-aligned motion sequences using multimodal foundation models. The inference process follows a three-stage procedure designed to ensure broad adaptability by decoupling high-level sequencing from specific robotic constraints: First, using $\mathcal{C}$, the model is prompted to generate a contextually appropriate action sequence. The sequence is then mapped against the robot’s capabilities to verify execution feasibility, and finally, the system queries the skill library to synthesize motion behavior that best aligns with the social context under the robot's kinematic constraints.
                
        Let $\mathcal{S}$ be the scenario that consists of both the context description $\mathcal{C}$ and the robot-specific parameters $\mathcal{R}$ (see \autoref{fig:system_overview}). A robot's motion capability is denoted by $r_c \in \mathcal{R}$ and includes platform-specific affordance descriptions such as arm articulation or torso inclination. Each entry $r_y \in \mathcal{R}$ in the library of available motion primitives has three attributes: A unique name, a semantic description, and the motion primitive object itself.
        The output of this procedure is a behavioral plan $\pi$, representing a sequence of $N$ joint-space whole-body trajectories $y \in r_y$ where each $y_i$ is instantiated from the library. 

        $$\mathcal{S}=\{\mathcal{C}, \mathcal{R}\} \overset{\text{\shortstack{Sequence Generation}}}{\rightarrow} [y_1, \dots, y_N] \in\pi$$

    \subsection{Sequence Modulation} \label{sec:how}
        While the behavioral plan $\pi$ determines the sequence of actions (the ``what''), different contexts can necessitate meaningful variations in execution. In human-to-human interaction, for example, a greeting differs significantly depending on social factors, such as whether the individual was recently seen or if it is a long-awaited reunion. Although the underlying motion sequence may remain the same, its stylistic execution, or modulation, must vary to remain contextually appropriate (the ``how'').
        
        To this end, we build on past work \cite{hielscher2025interactive}, where we introduced Dynamic Movement Primitives (DMPs) to implement the Principles of Animation in a mathematically unified formulation. We briefly summarize the theoretical approach and operationalization of the individual principles, and then show how we extend it with automatic context-aligned parametrization (see \cite{hielscher2025interactive} or the supplementary material for more details).

        DMPs are multi-dimensional second-order dynamical systems, coupled by a phase variable $x$, oriented towards a goal state $g$ and encoding joint trajectories in a set of weights $w$ as part of the nonlinear forcing term $f(x)$:        
        $$\tau^2 \ddot{y} = \alpha \left( \beta \left( g - y \right) - \tau \dot{y} \right) + f(x).$$
        Locally weighted regression is used to learn the weights given a reference trajectory, which enables zero-shot imitation learning from demonstrated motion.
        
        Modulation of the principles is done by altering individual system components or parameters -- the acceleration $\ddot{y}$, the weights $w$, or the forcing term $f(x)$ -- to achieve the desired effects. These components are modified for each principle and per degree of freedom, allowing the creation of complex, multi-dimensional motion in a low-dimensional and interpretable parameter space defined by the Principles of Animation. Specifically, we consider each relevant principle and how we mathematically implement its individual modulation:
        
        \textit{Arcs} controls the smoothness of the motion and is modulated by convolving a 1D Gaussian kernel with the weights $w$ for $p_{\text{Arcs}} > 0$ to achieve smoother motion or by applying unsharp masking on the weights $w$ for $p_{\text{Arcs}} < 0$ to accentuate motion features.
        \textit{Anticipation} negatively scales the system's acceleration with $p_{\text{Ant}} > 0$ at the beginning of the trajectory to achieve a preceding anticipatory action.
        \textit{Slow In Slow Out} adapts the phase function to slow down system progression at the beginning and end of the trajectory for more natural motion.
        Specifically, we utilize an inverted sigmoid function to obtain the phase.
        \textit{Timing} adjusts the overall duration of the motion by scaling $\tau$ to achieve faster motion with $p_{\text{Time}} < 1$ and slower motion with $p_{\text{Time}} > 1$.
        \textit{Exaggeration} understates ($p_{\text{Exa}} < 1$) or overstates ($p_{\text{Exa}} > 1$) the motion features by scaling the effect of the forcing term $f(x)$.
        \textit{Follow Through} scales the values of a \textit{target joint} based on values of a \textit{source joint} with $p_{\text{Follow}} > 0$ to create an effect of connected bodies.
        In addition to these principles, the modulation of the goal $g$ enables target-oriented motion execution, e.g., toward a detected human or object in the environment. Furthermore, we introduce natural variability across executions by perturbing the weights $w$ with Gaussian noise scaled by $p_{\text{Rand}} > 0$, to generate less repetitive motion.
        
        The representation of expressivity in the Principles of Animation, first formalized by Thomas and Johnston \cite{thomas1995illusion}, provides a powerful framework for expressive character animation, and as just shown, for parameterizing motion using DMPs. Importantly, this space is well understood and documented, and modern foundation models -- trained on extensive bodies of general knowledge as well as the specific literature on creating lifelike characters using the Principles of Animation -- can navigate it with minimal guidance. We therefore use foundation models to automatically generate the DMP modulation parameters that align motion execution to the given context. This process is split into three submodules, and \autoref{fig:system_overview} visualizes the connection of those aligned with the given transition formulas:
        
        The first module handles the semantic mapping from the context description $\mathcal{C}$ and the generated motion sequence $\pi$ to a style description of a motion execution. This step is analogous to a director guiding an actor's performance on a movie set. The output of this module, called modulation descriptions $\mathcal{M}_\mathbf{d}$, does not yet include a relation to the Principles of Animation but rather a natural language description of how the motion should be executed. For each motion primitive in the sequence, a description $\mathcal{M}_d$ is generated to fit the context description.        
        $$\{\pi, \mathcal{C}\} \overset{\text{\shortstack{Modulation Description}}}{\rightarrow} [\mathcal{M}_{d, 1}, \dots, \mathcal{M}_{d, N}]$$
        Notably, the motion descriptions are generated in an informed and coherent manner, where each primitive is described completely and independently, yet all descriptions remain consistent with the intent of the overall sequence. This is especially important with long sequences or sequences that include complex individual movements.
                
        The second module relates the given modulation descriptions $\mathcal{M}_\mathbf{d}$ to the Principles of Animation and generates ratings $\mathcal{M}_\mathbf{r}$ for all relevant principles. The module operates at the primitive instance level and, therefore, is called for every motion primitive within the sequence.        
        $$\mathcal{M}_{d, n} \overset{\text{\shortstack{Principle Relation}}}{\rightarrow} \mathcal{M}_{r, n},  \quad \forall n \leq N$$
        By utilizing the general knowledge of the foundation model, which includes the use of the Principles of Animation, the only guidance necessary is the explanation of a target rating scale for the individual parameters. Specifically, we generate individual ratings $\mathcal{M}_r \in \mathcal{M}_\mathbf{r}$ for the animation principles \textit{Anticipation}, \textit{Arcs}, \textit{Exaggeration}, \textit{Follow Through}, \textit{Slow In Slow Out}, and \textit{Timing}. All principles are rated using generic output scales, i.e., scales from 1 to 10 or scales from -5 to 5. The reason for using generic scales lies in better generalization and improved robustness.
                
        In the third module, the ratings from the generic scale $\mathcal{M}_r$ are projected on a robot-specific parameter range to create modulation parameters $\mathcal{M}_p$.       
        $$\{ \mathcal{M}_{r, n}, r_p\} \overset{\text{\shortstack{Parameter Projection}}}{\rightarrow} \mathcal{M}_{p, n},  \quad \forall n \leq N$$
        These parameter ranges need to be part of the robot-specific data $r_p \in \mathcal{R}$. The projection of the ratings for \textit{Anticipation} and \textit{Follow Through} is realized with linear mapping. The projection of the ratings for \textit{Arcs}, \textit{Exaggeration}, and \textit{Timing} is realized with piecewise linear mapping to provide an optimal relation to the scaling parameters of the DMP. \textit{Slow In Slow Out} can be either on or off.
                
        Finally, the adjustments to the motion execution are made by modulating individual DMP trajectories $y \in \pi$ with the modulation parameters $\mathcal{M}_p$ to obtain the modulated motion primitive trajectory $y_\mathcal{M}$ as part of the modulated sequence $\pi_\mathcal{M}$.
        $$y_n \overset{\text{Modulation}}{\rightarrow} y_{\mathcal{M}, n}, \quad \forall n \leq N$$
        
    \subsection{System Overview} \label{sec:framework}
        With the goal of autonomous and fluent interaction in real scenes, we propose a system for continuous and efficient operation using a hierarchical control structure based on a timeline. To decrease reaction times, we employ a knowledge-distilled multimodal foundation model tailored to generate expressive motion behavior with low inference times. Building on the methodological foundation above, we show how the system can be abstracted from specific robots to remain adaptable and flexible across varying capabilities, skills, and morphologies.
        
        \textbf{Timeline Control:}
            Using a timeline aligns naturally with the structure of the generated modulated sequences and leverages the inherent sequencing capabilities of DMPs. Transitions between two DMPs occur smoothly, as each DMP’s state is governed by a dynamical attractor system. Consequently, the timeline allows for continuous progression through the sequence by linking individual primitives to produce seamless motion.
            In practice, the timeline is implemented as a first-in-first-out (FIFO) queue, from which motion primitives are retrieved to determine the currently active primitive.
            For continuous actuation of the robot, the progressing state of the currently active primitive provides the joint control commands to move the robot.

            The hierarchical control strategy of our system operates at different frequencies.
            At a high level with lower frequency, the population of the timeline follows two governing cases: The first case is reactive to incoming interaction context $\mathcal{C}$ and adds the individual motion primitives of a generated modulated sequence $\pi_\mathcal{M}$ to the timeline for sequential execution.
            In this case, the timeline queue is first cleared before appending the new sequence, to allow for fast reaction. In an extreme case, and towards even faster reactions, the currently active primitive can be directly replaced, without finishing execution and directly transitioning to the first primitive of the new sequence.
            In the second case, during idle operation, idle motion primitives are appended and executed to ensure continuous control of the robot. The idle motion primitives are randomly selected from a separate library of idle gestures, which can be a subset of the primary gesture library $r_y$ or contain low-intent motion, or even static joint commands.
            At a lower level, and decoupled from the timeline population, the hierarchical control system enables continuous motion adjustments to modulate the executed motion based on relations between the DMP and real-time data. This modulation is executed at control frequency in an online fashion. When a detected person in the scene changes their location, the robot can adapt and maintain its orientation and gesture target during execution by updating the DMP goal state $g$. Notably, idle movements can also be continuously modulated to track a person or to induce random subtle movements using $p_\text{rand}$.
        
        \textbf{Knowledge Distillation and Generation Inference:}
            Since reaction time is critical for natural and fluent interaction, we optimize our generation method using knowledge distillation to fine-tune a small, efficient multimodal foundation model. This approach relies on supervised training with data from a larger model, which generally produces superior results and enables condensing multi-step inference pipelines into a single step. We leverage both aspects, i.e., our training set consists of traces generated by the multi-step pipeline using the large model. Specifically, our training data stems from diverse traces generated from a context dataset consisting of over 100 contextual descriptions $\mathcal{C}$ -- similar to the ones we present in the experiments section -- and specific data $\mathcal{R}$ for three different robots.
            The resulting fine-tuned foundation model takes the same inputs and produces the same outputs as the multi-step pipeline, but consumes fewer input and output tokens, as intermediate steps are cut.            
            Generally, and this applies to both the single call of the fine-tuned model as well as the multi-step pipeline, we utilize structured I/O interfaces to ensure that the inference output adheres to the required form, which is especially important with respect to system robustness.
        
        \textbf{Cross-Morphology:}
            The system is designed to be robot agnostic, enabling its use across different platforms. To operate, the system requires the following inputs: \textit{Capabilities}: A textual description of the robot’s motion-related abilities. For example, a human-like robot can move its head, arms, and torso, while a quadruped can move its legs and torso. \textit{Kinematics}: The robot’s kinematic structure, provided as a URDF file. \textit{Movements}: A collection of existing joint trajectories for gestures or other task-specific motions and interaction skills. \textit{Parameter Limits}: Bounds for individual modulation parameters. While defaults are often sufficient, the limits can be customized to fine-tune modulation.
            
            Based on these inputs, a preprocessing module computes robot-specific data $\mathcal{R}$, necessary for system operation, in two steps: First, joint trajectories are encoded to DMPs, forming the motion primitive gesture library. Second, the URDF file is analyzed to construct a kinematic map and compute the global axes of each joint. This information allows the preprocessing module to identify valid source-target joint pairs for the \textit{Follow Through} principle, which must reside within the same kinematic chain and be either directly connected or separated by a single condition joint. Axis alignment within the chain is verified based on the configuration of the condition joint.

%% file: sections/experiments.tex
\section{Experiments}
    The evaluation of our method is carried out in multiple experiments that address different aspects of the system.
    Specifically, we conduct a user study to investigate the subjective perception of context alignment of the generated motion behavior and compare our method against an API-level baseline condition.
    Additionally, we analyze the inference times of the generation process to validate that our approach of knowledge distillation achieves a cut in system reaction times.
    Further experiments aim to show the handling of multimodal context descriptions as well as the integration of robots with different kinematic structures into our system.

    In our experiments, we use the following generation conditions: \textit{Ours}: Our system, which automatically generates and modulates gesture sequences. The multimodal foundation model is a version of OpenAI's gpt-4.1-nano, which is learned with knowledge distillation from \textit{Ours (multi)} as introduced below. \textit{Ours (non-mod)}: An ablation of our system, which automatically generates gesture sequences that are not modulated. \textit{API-level}: A baseline condition that automatically generates behavior programs from an API library of the robot. This is a reimplementation of the method proposed in \cite{mahadevan2024generative}. \textit{Ours (multi)}: Our method but with the multi-step pipeline, which interfaces gpt-4.1.
    
    For the majority of our experiments, we deploy the \textit{Pepper} robot, with a motion primitive library of over 30 complex and expressive whole-body gestures, for e.g., waving, pointing, or explaining.
    
    We use the following context descriptions for different scenarios between the robot and an interaction partner:
    \begin{description}
        \item[\textit{Context 1}]\textit{``A person approaches and greets the robot, then asks for directions and an explanation. The robot knows that the destination is to its left. The robot's energy level is very low.''}
        \item[\textit{Context 2}]\textit{``A person familiar to the robot approaches from its right side. The robot hasn’t seen this person in a long time, and the person is clearly excited to meet the robot again.''}
        \item[\textit{Context 3}]\textit{``A person walks past the robot. The person is busy and in a hurry.'' }
        \item[\textit{Context 4}]Audio modality: The sound of a person ``[knocking]'' on a door; Textual modality: \textit{``The robot is engaged in a discussion with a person. It is also anticipating the arrival of a second person. An open door is located to the left.''}
    \end{description}

    We conducted a user study, for which we recruited 38 participants - 26 male, 12 female - aged between 21 and 84 from mailing lists and university information boards. 8 participants did not finish the survey, and their partially filled questionnaires are not regarded in the evaluation. The study consists of multiple scenarios, which appear in randomized order and are divided into two questionnaire parts. In both, participants watch videos of the human-like robot \textit{Pepper} and answer questions related to the videos. All videos show both the robot and the silhouette of the robot's interaction partner.
    
    \textbf{Study - Experiment 1:}
        For the first part of the study, we generate expressive robot motion behavior using the conditions \textit{Ours}, \textit{Ours (non-mod)}, and \textit{API-level}.
        The \textit{API-level} condition is provided with a library of API calls to move the arms, the head, and the torso of the robot, as well as functions for scene understanding, like getting the yaw and pitch angles of humans in the scene.
    
        The questionnaire follows a within-subject structure, where videos of all three conditions appear in randomized order. For each video, the participants rate how well the shown motion behavior aligns with the context that describes the scenario on a 7-point linear scale.
        With this experiment, we study the hypothesis \textbf{H1:} \textit{The perceived alignment between context and expressivity is rated higher for non-modulated sequences than for API-level behaviors}, as well as the hypothesis \textbf{H2:} \textit{The perceived alignment between context and expressivity is rated higher for modulated sequences than for non-modulated sequences}.
    
    \textbf{Study - Experiment 2:}
        In the second part of the study, we compare the generated expressive motion behavior from the two conditions \textit{Ours} and \textit{Ours (multi)}.

        The questionnaire shows videos of both conditions in randomized order, and participants are asked to rate which of the two conditions aligns better with the given context description. A continuous bipolar slider ranging from -100 to 100 is used. The extremes (-100 and 100) indicate that one of the two conditions is perceived as much better aligned, while the center (0) marks equal alignment. This scale allows preferences to be directly projected onto a percentage scale, providing an intuitive measure of relative alignment.
        Towards verifying that the efficiency enhancements as part of our method do not diminish the quality of generated expressive motion behavior, we study the hypothesis \textbf{H3:}\textit{The perceived alignment between context and expressivity for the knowledge-distillation model is not significantly inferior to that of the multi-step pipeline, given the specified non-inferiority margin.}
    
    \textbf{Experiment 3:}
        Related to the previous experiment, we compute inference times, and the thereby resulting system latency, for both conditions \textit{Ours} and \textit{Ours (multi)}.
        For \textit{Context 1} to \textit{Context 3}, we time the generation process over 100 runs per context. This results in a total of $3 \times 100$ inference calls for Ours and $3 \times (2 \times 100 + N \times 100)$ calls for \textit{Ours (multi)}, where $N$ is the variable sequence length.
    
    \textbf{Experiment 4:}
        We deploy our framework on different robots to support the claim that our method is decoupled from specific robots.
        To this end, we use the following robots:
        \textit{Aldebaran Pepper}: A mobile robot with a human-like morphology that can move its torso, head, and arms with 17 degrees of freedom.
        \textit{Unitree Go2}: A quadruped platform with 12 degrees of freedom.
        \textit{Kinova Gen3 Lite}: An industrial robot arm with 6 degrees of freedom.
        The decision for this selection lies in the fundamentally different morphologies and kinematic structures of the listed robots.
        We generate motion behavior using our method for the scenario described with \textit{Context 2} for all three robots.
    
    \textbf{Experiment 5:}
        We want to show how our system handles multimodal context descriptions and generates motion behavior for the following two scenarios:
        First, \textit{Context 4}, where a textual context description, detailing internal states and the knowledge of the robot, is augmented by an audible knocking sound.
        Second, a multimodal version of \textit{Context 2}, where the video and audio modalities are used directly rather than transcribed into text. Specifically, for the video modality, it is the image of a person who appears happy, and the audio modality is given in the form of spoken language -``Hi! It's amazing to see you. It's been so long''. The textual modality describes internal knowledge of the robot.

%% file: sections/results.tex
\section{Results and Discussion}
    \begin{figure*}[t]
        \centering
        \includegraphics[width=0.99\linewidth, alt={An overview figure for the quantitative results of the experiments, which is split into three parts and visualizes the results given in the results sections.}]{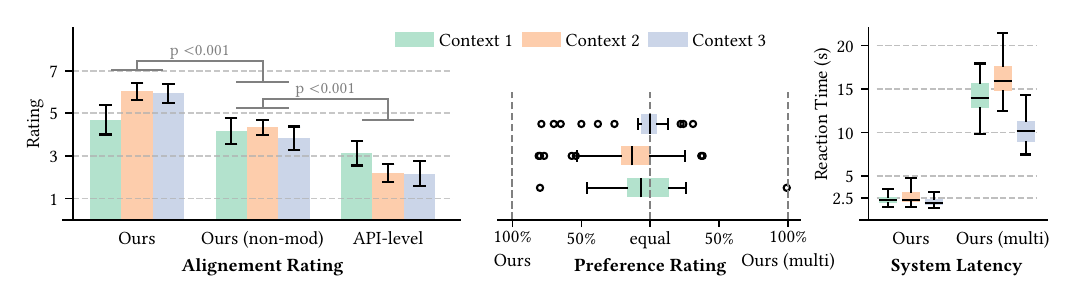}
        \vspace{-6mm}
        \caption{\textbf{Quantitative results: The alignment ratings from Experiment 1 (left) show a significant difference between context alignment of the different generation conditions. The perceived preference ratings from Experiment 2 (center), with a slight shift towards \textit{Ours}, are statistically non-inferior with a non-inferiority bound of 10 percent. The system latency comparison from Experiment 3 (right) shows a significant reaction time reduction for the distilled model.}}
        \Description{An overview figure for the quantitative results of the experiments, which is split into three parts and visualizes the results given in the results sections.}
        \label{fig:results}
    \end{figure*}
    \begin{figure*}[t]
        \centering
        \includegraphics[width=0.99\linewidth, trim={0 84px 0 0 }, clip, alt={Sequences of images showcasing the executed motion behavior for Pepper, Go2 and Gen3 Lite.}]{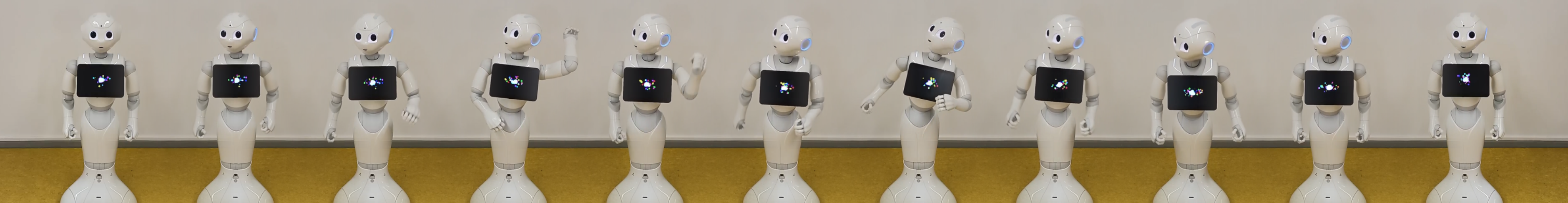}
        \includegraphics[width=0.99\linewidth, trim={0 57px 0 27px}, clip, alt={Sequences of images showcasing the executed motion behavior for Pepper, Go2 and Gen3 Lite.}]{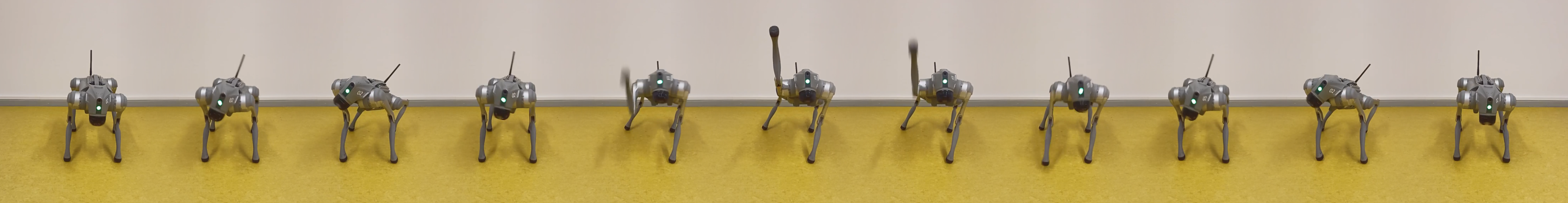}
        \includegraphics[width=0.99\linewidth, trim={0 42px 0 42px}, clip, alt={Sequences of images showcasing the executed motion behavior for Pepper, Go2 and Gen3 Lite.}]{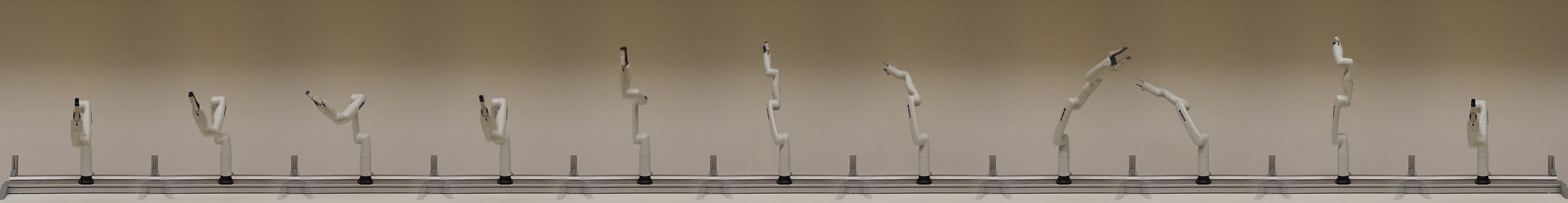}
        \caption{\textbf{Motion behavior for Context 2 generated with our system for \textit{Pepper} (top), \textit{Go2} (center), and \textit{Gen3 Lite} (bottom).}}
        \Description{Sequences of images showcasing the executed motion behavior for Pepper, Go2 and Gen3 Lite.}
        \label{fig:sequences}
        \vspace{-3mm}
    \end{figure*}

    \subsection{Quantitative Results}
    
    \textbf{Results - Study - Experiment 1:}
        Data is analyzed using a two-way repeated-measures ANOVA to assess the effects of experimental factors on the collected ratings. Following the significant effect for the generation conditions ($p < 0.001 $), paired-samples t-tests, with Bonferroni correction applied, are used for specific comparisons to investigate our hypotheses. The results of these post-hoc tests show a significant difference between \textit{Ours} and \textit{Ours (non-mod)} ($p < 0.001$), and between \textit{Ours (non-mod)} and \textit{API-level} ($p < 0.001$). The results are visualized on the left in \autoref{fig:results}. Our hypotheses \textbf{H1} and \textbf{H2} are supported.
    
    \textbf{Results - Study - Experiment 2:}
        To avoid false statements based on averaging factors, we conduct per-context non-inferiority tests using a one-sided t-test with a non-inferiority margin of \SI{10}{\%} of the slider range. The results indicate significant non-inferiority of the generation condition \textit{Ours} compared to \textit{Ours (multi)} with strong statistical evidence ($p<0.01$) for \textit{Context 1} and very strong statistical evidence ($p<0.001$) for \textit{Context 2} and \textit{Context 3}. The respective rating means are \SI{3.86}{\%}, \SI{16.73}{\%}, and \SI{7.56}{\%} in favor of \textit{Ours}. The results are visualized in the center of \autoref{fig:results}. Our hypothesis \textbf{H3} is supported.
    
    \textbf{Results - Experiment 3:}
        We evaluated system reaction times across \textit{Context 1–3}. The knowledge distillation model (\textit{Ours}) averaged \SI{2.40}{s}, \SI{2.66}{s}, and \SI{2.22}{s} per context respectively, with an overall mean of \SI{2.39}{s} using an average of 2161 tokens. 
        The multi-step pipeline (\textit{Ours (multi)}) averaged \SI{14.32}{s}, \SI{16.66}{s}, and \SI{10.26}{s}, with an overall mean of \SI{12.87}{s} using an average of 4969 tokens.
        This corresponds to a \SI{5.4}{\times} speedup, or an \SI{81.5}{\%} reduction in reaction time, demonstrating substantially lower system latency with the knowledge distilled model. The results are visualized on the right in \autoref{fig:results}.
    
    \subsection{Qualitative Results}
    \textbf{Results - Experiment 4:}
        The generated motion behavior for all three robots is visualized in \autoref{fig:sequences}. All robots direct their movements towards the side of the approaching person, either based on selection of gestures (\textit{Go2, Gen3 Lite}) or the modulation of $g$ (\textit{Pepper}, \textit{Gen3 Lite}). The sequences include greeting gestures modulated to match the described enthusiasm of the approaching person.
    
    \textbf{Results - Experiment 5:}
        For \textit{Context 4}, the robot first greets and then invites the person closer. Without the audio modality, however, our method only produces explaining and elaborating gestures, showing that the knocking sound is correctly interpreted and the generated motion aligns better with what happens in the scene (see \autoref{fig:title}).
        For the multimodal version of \textit{Context 2}, including an image of a person who appears happy together with the voice recording, the generated motion closely matches the one solely based on the text description, on both levels, the expressivity and the generated parameter values. This confirms that multimodal inputs that match or reinforce the context produce the same resulting motion.

    \subsection{Discussion}

        The results of our experiments show how our method enables the context-aware generation and modulation of expressive motion behavior both quantitatively and qualitatively. Specifically, the evaluation of our user study supports all our hypotheses with strong ($p<0.01$) or very strong ($p<0.001$) statistical evidence. The evaluation, while not showing an effect on the ratings from context alone, does show statistical evidence ($p<0.05$) of an effect of the context on the generation conditions. However, the effect size is quite small with $\eta^2 = 0.0918$ compared to that of the generation condition alone ($\eta^2 = 0.447$, $p<<0.001$), but can still be seen in the plotted alignment rating, especially for \textit{Context 1}.
        These findings can possibly be grounded in individual perception from participants that reached out using our contact information, stating comments like the expectation of a robot to still function normally even if its energy reserves are low (\textit{Context 1}), or embodiment problems with lacking hand opening or closing, which is not possible with our application of the \textit{Pepper} robot.
        Also notable is that the collected preference ratings actually show a slight preference towards the knowledge distilled model. We do not have a clear explanation for this, but speculate that it may be related to the model’s ability to capture the overall context more effectively -- without relying on intermediate steps -- while still managing high complexity.
        The analysis of the latency shows the anticipated improvement of system reaction time in favor of a more fluent interaction \cite{shiwa2008quickly}.
        Another relevant aspect is the importance of extensive libraries for the optimal utilization of our method. Our gesture library for \textit{Pepper} is rather large, with over 30 different basic gestures, and our method produced diverse sequences. However, we noticed repeating patterns on the Go2 and Gen3 Lite, which use libraries that are less exhaustive, but still consist of whole-body gestures.
        The results of our experiment to test multimodal context descriptions suggest that incorporating video and audio modalities directly does not compromise the expressivity, and even allows us to condition the generation on audio cues that are not spoken language.

%% file: sections/conclusion.tex
\section{Conclusion}
    In this work, we present a system to automatically generate robot motion behavior from a multimodal description of an interaction context.
    Using multimodal foundation models, our method sequences and modulates whole-body gestures, encoded as Dynamic Movement Primitives, to align the robot's expressivity to the context.
    With the integration into a timeline-based continuous controller that leverages knowledge distillation and is adaptable to robots with different kinematics, the approach extends the state of the art by providing more generality, interpretability, efficiency, and flexibility.
    In a user study, participants rated the motions produced by our system as better aligned with the interaction context than those generated under a non-modulated or API-level motion condition. Further, participants did not perceive a difference in context alignment when using the distilled model, which has a system latency reduced by \SI{81.5}{\%} on average. Our qualitative experiments show the adaptability of the system to different robots as well as the handling of multimodal context inputs, including images, text, and sounds. Future extensions could include higher-level cognitive components, such as an episodic memory or a theory-of-mind model, to integrate continuous perceptions into a coherent and persistent context representation, as well as extending the system to verbal outputs combined with synchronized co-speech gestures.